\documentclass[10pt,twocolumn,letterpaper]{article}

\usepackage{cvpr}      
\usepackage{array} 
\usepackage{multirow}
\usepackage{booktabs}
\usepackage[table]{xcolor}
\usepackage{colortbl}
\usepackage{xcolor}

\usepackage{cuted} 

\usepackage{currfile} 

\usepackage{caption} 
\definecolor{cvprblue}{rgb}{0.21,0.49,0.74}
\usepackage[pagebackref,breaklinks,colorlinks,allcolors=cvprblue]{hyperref}

\def\confName{CVPR}
\def\confYear{2027}

\title{CreaTrace: Video Creative Plagiarism Detection and Evidence Tracing}

\author{
Chang Liu$^{1}$ \quad
Sheng Li$^{1}$ \quad
Zhenxing Qian$^{1}$ \quad
Xinpeng Zhang$^{1,*}$\\
$^{1}$Fudan University\\
{\tt\small 26113050283@m.fudan.edu.cn,
\{lisheng, zxqian, zhangxinpeng\}@fudan.edu.cn}\\}

\begin{document}
\maketitle
\begin{strip}
\centering
\includegraphics[width=1.0\linewidth]{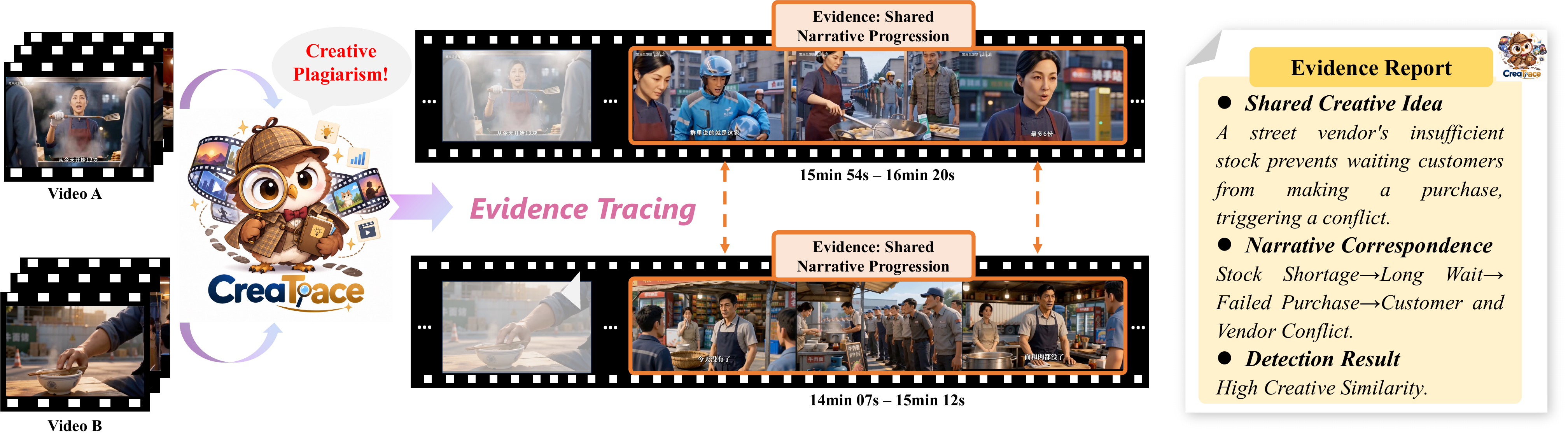}\hfill
\captionsetup{hypcap=false}
\captionof{figure}{\textbf{Illustration of video creative plagiarism detection and evidence tracing by our CreaTrace. } Given two videos which share a similar creative idea but with different visual appearance, our CreaTrace identifies and localizes the corresponding evidence segments in both videos and generates an interpretable report highlighting their shared creative content and narrative structures.}
\captionsetup{hypcap=true}
\label{first}
\end{strip}
\begin{abstract}
Video creative plagiarism refers to the unauthorized reproduction of creative ideas across videos despite substantial differences in visual appearance, posing a new challenge to video copyright protection. However, existing video copy detection methods primarily focus on visual content similarity, struggling to capture abstract creative that are entangled with specific visual realizations. To address this issue, we introduce a novel task, named \textbf{Video Creative Plagiarism Detection (VCPD)}, and propose an iterative creative evidence tracing framework, termed \textbf{CreaTrace}, to verify creative similarity and accurately localize the corresponding evidence segments. Specifically, we design a Multimodal Semantic Event Segmentation strategy to divide similar long videos into coherent narrative events, facilitating event-level dataset construction and retrieving similar segments. Based on these events, CreaTrace employs Coarse Creative Matching to identify candidate event pairs through narrative semantics and camera language. Then CreaTrace presents Creative Decoupling to disentangle abstract creative patterns from specific visual realizations. Meanwhile, Creative Swap Contrastive Learning is devised to enhance creative consistency through cross-realization feature swapping. Benefiting from the above designs, our CreaTrace enables accurate creative similarity detection and temporal localization, while providing interpretable evidence report. Extensive experiments demonstrate the effectiveness and robustness of CreaTrace across diverse creative plagiarism scenarios.
\end{abstract}    
\section{Introduction}
\label{Introduction}

\begin{figure}
  \centering
   \includegraphics[width=1.0\linewidth]{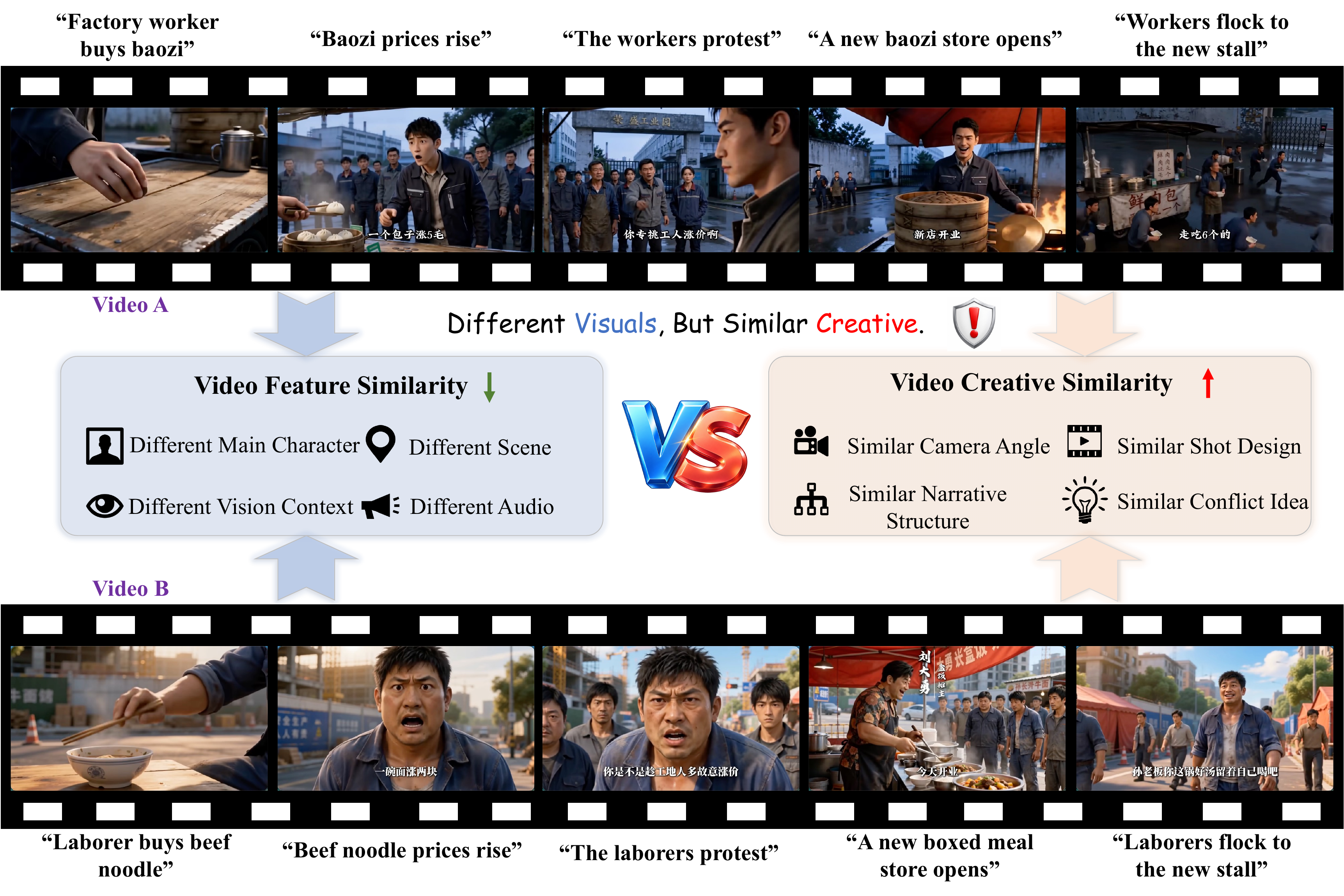}
   \caption{\textbf{Motivation.} Two creative plagiarized videos differ substantially in visual content but share similar underlying creative designs, including narrative structures, shot designs and conflict.}
   \label{motivation}
\end{figure}

The rapid development of short-video platforms and video generation models \cite{sora,wan2025wan,yang2025cogvideox} have made video creation more convenient. However, these advances have also raised concerns about identifying creative plagiarism and protecting original video copyright. Unlike traditional video copying \cite{ssan,he2023transvcl,wang2026mlt} which directly reuses visual content, creative plagiarism reproduces the creative design, such as narrative structure \cite{chun2024aistorysimilarity} and shot arrangements \cite{shot}, while changing its visual appearance. As illustrated in Figure \ref{motivation}, two videos differ in characters and scenes but share similar storylines and underlying creative designs, which poses a new challenge to video copyright protection.\par

Existing studies on video similarity mainly focus on duplicate video retrieval \cite{kordopatis2019visil}, video copy detection and localization \cite{vcsl,he2023transvcl}. Representative methods typically rely on visual feature matching and spatio-temporal correspondences to identify similar video segments. However, their reliance on visual feature similarity makes it struggling to detect creative plagiarism because similar creative designs can be expressed through different visual content. Therefore, detecting creative plagiarism requires abstracting intrinsic creative patterns by disentangling creative designs from their special visual realizations.\par

Recent advances in multimodal large language models \cite{bai2025qwen3,lin2024video,yan2026videochat} have significantly improved video perception and reasoning \cite{qiu2025videosetdiff}, further improving video understanding. However, video semantics primarily describe what occurs in a video, whereas creative similarity focuses on how events are organized, developed and visually presented. For example, videos with different semantic content may share similar narrative structures and shot arrangements. Consequently, directly comparing semantic representations remains insufficient for creative plagiarism detection.\par

In this paper, we focus on a new task named \textbf{Video Creative Plagiarism Detection (VCPD)}, which aims to identify potential creative plagiarism between videos with different visual appearances but similar creative structures. Unlike traditional video copy detection, VCPD focuses on how video content is organized and presented rather than whether the same visual elements appear in both videos. These abstract creative patterns are often entangled with specific visual realizations, making it challenging to identify creative similarity across visually distinct videos.\par

To address these challenges, we introduce a Multimodal Semantic Event Segmentation strategy to divide long videos into coherent narrative events, facilitating event-level dataset construction and creative similarity analysis. Building on this, we propose \textbf{CreaTrace}, an iterative creative evidence tracing framework for creative plagiarism detection and evidence localization. Specifically, CreaTrace first performs Coarse Creative Matching to retrieve candidate event pairs based on narrative semantics and camera language. Subsequently, CreaTrace employs Creative Decoupling to separate abstract creative patterns from specific visual realizations, while Creative Swap Contrastive Learning further enhances creative representation consistency by cross-realization feature swapping. Guided by the learned creative representations, CreaTrace iteratively localizes creative evidence in the original videos and feeds the localized evidence back to refine creative matching, progressively improving similarity verification and temporal localization. As shown in Figure~\ref{first}, our CreaTrace can detect creative plagiarism, localize supporting evidence segments and generate interpretable reports. Extensive experiments demonstrate the effectiveness of our CreaTrace in creative plagiarism detection and evidence localization.\par
Our contributions are summarized as follows:
\begin{itemize}
    \item We propose a novel task, named Video Creative Plagiarism Detection (VCPD) which aims to identify shared creative patterns across visually different videos.
    \item We design a Multimodal Semantic Event Segmentation (MSES) strategy to divide long videos into coherent narrative events, facilitating event-level dataset construction and fine-grained detection and localization.
    \item We propose CreaTrace, an iterative creative evidence tracing framework that integrates Creative Decoupling and Creative Swap Contrastive Learning to learn disentangled creative representations, and iteratively refines creative matching through localized evidence feedback.
    \item Extensive experiments demonstrate our CreaTrace enables precise creative plagiarism detection and temporal localization, while providing interpretable reports.
\end{itemize}

\begin{figure*}[t]
  \centering
   \includegraphics[width=1.0\linewidth]{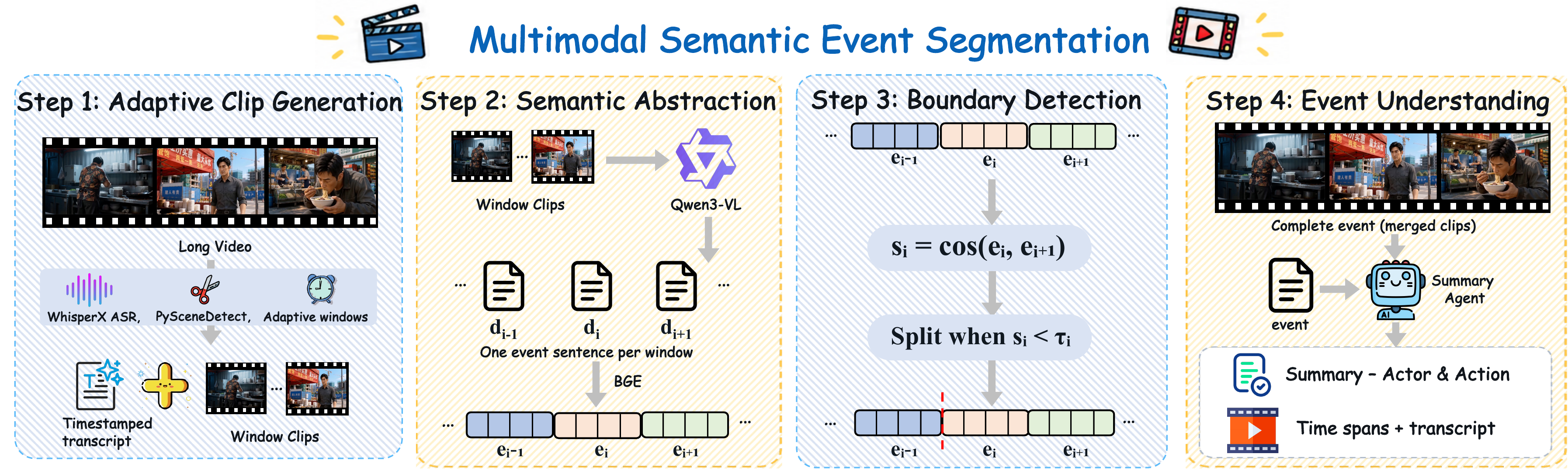}
   \vspace{-0.5cm}
   \caption{\textbf{Overview of MSES.} MSES consists of four stages: adaptive clip generation, semantic abstraction, boundary detection and event understanding. By integrating multimodal cues and semantic similarity, MSES segments long videos into coherent events and extracts structured information, including event summaries, actors, actions and temporal boundaries, to support event-level dataset construction.}
   \label{mses}
\end{figure*}

\section{Related Work}

\subsection{Video Similarity Learning and Retrieval}
Video similarity learning aims to quantify the video content correspondence between a pair of videos, supporting various downstream applications such as video retrieval and copyright protection. Early methods typically employ global video representations for efficient retrieval but often overlook spatio-temporal correspondences \cite{jiang2014vcdb,kordopatis2019fivr}. To address this limitation, ViSiL \cite{kordopatis2019visil} models spatio-temporal correspondences through frame-level similarity matrices, improving video retrieval accuracy. While DnS \cite{kordopatis2022dns} distills fine-grained similarity knowledge into lightweight student models, achieving a better balance between retrieval accuracy and computational efficiency. Subsequent studies further explore self-supervised similarity learning \cite{svs} and informative frame selection \cite{jo2024vvs} to improve video representations. More recently, video similarity research has expanded beyond visual content matching. Specifically, ConViS-Bench \cite{liberatori2026convis} introduces concept-based similarity annotations for interpretable video comparison, while VideoSetDiff \cite{qiu2025videosetdiff} explores cross-video reasoning through explicit similarity and difference analysis. Additionally, SemanticMoments \cite{huberman2026semanticmoments} propose temporal statistics to explore motion similarity, reducing reliance on visual appearance. Despite these advances, above methods primarily focus on visual content similarity. In contrast, our work aims to identify creative similarity across different visual realizations.

\subsection{Video Copy Localization}

Video copy localization aims to identify corresponding temporal segments containing reused content between videos. Early methods typically construct frame-level similarity matrices and perform temporal alignment to locate copied segments. Specifically, SSAN \cite{ssan} combines keyframe selection with similarity pattern detection for efficient segment localization. While TransVCL \cite{he2023transvcl} further employs attention-based feature enhancement and temporal alignment for copy localization. Subsequent studies have explored more effective representations  and matching strategies. For example, iESTA \cite{ding2024iesta} combines instance-level spatial features with global temporal information to improve cross-video alignment. Meanwhile, FiGVCL \cite{luo2025figvcl} leverages unsupervised local feature matching to identify corresponding segments. More recently, MLT-Dedup \cite{wang2026mlt} combines clip-level retrieval with fine-grained frame matching to localize duplicated video segments. However, existing methods mainly focus on locating visually copied or modified content, whereas our work aims to identify and localize creative plagiarism segments.\par

\subsection{Multimodal Video Understanding Models}

Recent advances in multimodal video understanding models have significantly improved video content perception and reasoning. VideoChat2 \cite{li2025videochat} and LLaVA-Video \cite{zhang2024llava} enhance video-language understanding through multimodal instruction tuning. Subsequently, VideoLLaMA3 \cite{zhang2025videollama} further strengthens visual perception through vision-centric learning. Beyond semantic understanding, TimeChat \cite{ren2024timechat} introduces timestamp-aware modeling for temporal grounding, while InternVideo2.5 \cite{wang2025internvideo2} explores long-range temporal dependency through hierarchical token compression. More recently, Qwen3-VL \cite{bai2025qwen3} introduces interleaved spatio-temporal positional encoding and timestamp alignment for long video understanding. Meanwhile, InternVL3 \cite{zhu2025internvl3} employs native multimodal pretraining and variable visual position encoding to support extended multimodal contexts. Building on these advances, our work leverages multimodal video understanding to extract event-level semantic information, providing a foundation for subsequent creative matching and iterative evidence tracing.
\begin{figure*}[t]
  \centering
   \includegraphics[width=1.0\linewidth]{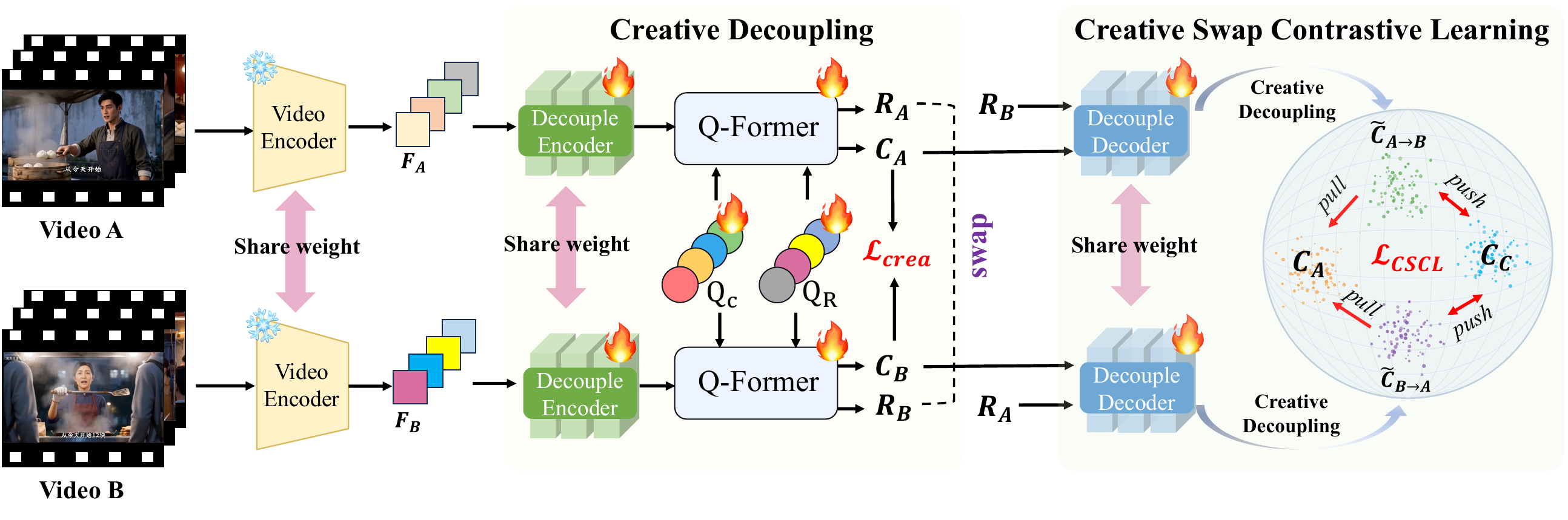}
   \vspace{-0.5cm}
   \caption{\textbf{Overview of our CreaTrace.} We propose two key designs to learn creative representations: Creative Decoupling and Creative Swap Contrastive Learning. The former disentangles creativity from specific video features through learnable queries, while the latter exchanges realization features and leverages contrastive learning to enhance creative consistency. Benefiting from above designs, our CreaTrace can effectively capture shared creative patterns for evidence localization.}
   \label{main}
\end{figure*}

\section{Methodology}

\subsection{Multimodal Semantic Event Segmentation}

Videos often comprise multiple narrative events, making it challenging to identify and localize potential creative plagiarism directly from entire videos. Meanwhile, shot-based segmentation may fragment semantically coherent events. To address these issues, we introduce Multimodal Semantic Event Segmentation (MSES), which integrates visual cues, speech transcripts and shot transitions to segment videos into coherent narrative events. As illustrated in Figure~\ref{mses}, MSES consists of four stages: adaptive clip generation, semantic abstraction, boundary detection and event understanding. We will detail each stage as follows.\par

\noindent\textbf{Adaptive Clip Generation.}
Given a video $V$ of duration $T$, we extract timestamped speech transcripts using WhisperX \cite{bain2023whisperx} and shot boundaries using PySceneDetect. To capture local semantics, we divide the video into non-overlapping analysis windows with an adaptive duration defined as:
\begin{equation}
\Delta T=\operatorname{Clip}\left(
\Delta_{\mathrm{ref}}\sqrt{\frac{T}{T_{\mathrm{ref}}}},
\Delta_{\min},\Delta_{\max}
\right),
\end{equation}
where $T_{\mathrm{ref}}$ and $\Delta_{\mathrm{ref}}$ denote the reference duration and window size, respectively. After rounding $\Delta T$ to a predefined interval, we obtain $N$ non-overlapping windows $\mathcal{U}=\{u_i\}_{i=1}^{N}$ with transcripts $\{l_i\}_{i=1}^{N}$.

\noindent\textbf{Semantic Abstraction.}
For each window $u_i$, we employ Qwen3-VL \cite{bai2025qwen3} to generate an event description $d_i$ from its visual content and aligned transcript $l_i$, capturing key actions, interactions and narrative developments. The description is then encoded into a normalized semantic representation using a pretrained BGE encoder \cite{bge}:
\begin{equation}
d_i=\mathcal{F}_{\mathrm{Qwen}}(u_i,l_i),\quad
e_i=\operatorname{Norm}\left(\mathcal{F}_{\mathrm{BGE}}(d_i)\right),
\end{equation}
where $e_i$ denotes the semantic feature of $u_i$, forming a feature sequence $\mathcal{E}=\{e_i\}_{i=1}^{N}$ for boundary detection.

\noindent\textbf{Boundary Detection.}
To identify event boundary, we compute the cosine similarity between consecutive semantic features and incorporate shot transitions to adaptively adjust the boundary threshold, which can be expressed as:
\begin{equation}
s_i=\cos(e_i,e_{i+1}),\quad
\tau_i=\tau_{\mathrm{sem}}+\lambda_{\mathrm{shot}}c_i,
\end{equation}
where $\tau_{\mathrm{sem}}$ denotes the base semantic threshold, $\lambda_{\mathrm{shot}}$ controls the influence of shot transitions and $c_i\in{0,1}$ indicates the presence of a nearby shot boundary. An event boundary is detected between adjacent windows when $s_i<\tau_i$.\par

\noindent \textbf{Event Understanding. }Based on the detected boundaries, consecutive windows are grouped into temporally ordered events $\left \{ {E_j} \right \} _{j=1}^{M}$, where $E_j=[t_j^s:t_j^e]$ denotes the temporal interval of the $j$-th event. Short segments are merged with preceding events to reduce fragmentation. For each event, Qwen3-VL analyzes its visual content and transcript to generate a structured description covering the event summary, actors, action sequence, narrative role and visual strategy. These event-level representations serve as the basis for subsequent creative matching and evidence tracing.\par

\subsection{Coarse Creative Matching}
Given the semantic events constructed by MSES, we perform coarse creative matching from two complementary perspectives: event semantics and camera language. Specifically, we compute semantic similarity $s_{ij}^{\mathrm{sem}}$ using cosine similarity between BGE embeddings of event summaries and action sequences, while camera language similarity $s_{ij}^{\mathrm{cam}}$ is obtained by aligning shot-level descriptors through Dynamic Time Warping \cite{dtw}, considering camera techniques, editing rhythms and shot grammar. Based on these scores, our CreaTrace identifies suspicious event pairs and further localize potential creative evidence, progressively focusing on highly suspicious segments. More details about coarse creative matching are supported in \textbf{Supplementary Material}.





\subsection{CreaTrace}

Given a set of suspicious video event pairs retrieved by Coarse Creative Matching, CreaTrace aims to verify their creative similarity and localize supporting evidence in the original videos. As illustrated in Figure \ref{main}, CreaTrace consists of two main components: (1) Creative Decoupling, which disentangles abstract creative patterns from specific visual realizations; (2) Creative Swap Contrastive Learning, which enhances the consistency of creative representations through cross-realization feature swapping. Building upon these representations, iterative Creative Evidence Tracing progressively localizes suspicious segments and feeds the retrieved evidence back to refine creative matching and guide subsequent localization. We will introduce each key components in detail as follows.\par

\noindent \textbf{Creative Decoupling.} Two videos sharing an underlying creative idea may exhibit substantial differences in characters, scenes and visual appearances. To distinguish intrinsic creative patterns from such variations, we introduce Creative Decoupling, which decomposes video features into creative and realization representations.

Specifically, given two video segments $V^A$ and $V^B$, we extract their temporal features using a shared frozen video encoder $\mathcal E_v$ and employ a trainable decoupling encoder $\mathcal E$ to map the extracted features into a latent space, as follows:

\begin{eqnarray}
F_i=\mathcal E_v(V_i), Z_i=\mathcal E(F_i),
\end{eqnarray}
\noindent where $i\in{A,B}$, $F_i$ denotes the extracted video features and $Z_i$ represents the latent features.

To disentangle creative from visual realization, we employ a dual-query Q-Former with learnable creative and realization queries, $Q_C$ and $Q_R$, respectively. The two queries independently attend to the latent features to extract their corresponding representations, which is defined as:
\begin{eqnarray}
C_i = Q_\theta(Q_C,Z_i),\
\end{eqnarray}
\begin{eqnarray}
R_i = Q_\theta(Q_R,Z_i).
\end{eqnarray}

To encourage videos sharing the same underlying idea to exhibit consistent creative representations despite different visual realizations, we design a creative consistency loss to explicitly align the creative representations, as follows:

\begin{eqnarray}
\mathcal{L} _{crea} = \frac{1}{K_c}|C_A-C_B|^2,
\end{eqnarray}
where $K_c$ denotes the number of creative queries.



\noindent\textbf{Creative Swap Contrastive Learning. }To further enhance creative discriminability across different visual realizations, we introduce a creative swap contrastive learning strategy. Specifically, given a pair of videos sharing the same underlying creative idea $(C_A,R_A)$ and $(C_B,R_B)$, we exchange their realization representations while preserving the creative components. The swapped features are reconstructed using a shared decoupling decoder $\mathcal D$, as follows:

\begin{eqnarray}
\widetilde F_{A\rightarrow B}&=\mathcal D(C_A,R_B),\\
\widetilde F_{B\rightarrow A}&=\mathcal D(C_B,R_A).
\end{eqnarray}

Then we perform creative decoupling on the swapped features $\widetilde F_{A\rightarrow B}$ and $\widetilde F_{B\rightarrow A}$ to obtain $\widetilde C_{A\rightarrow B}$ and $\widetilde C_{B\rightarrow A}$. To preserve the shared creative identity while enhancing inter-idea discrimination, we align the swapped creative representations with both $C_A$ and $C_B$, while contrasting them against the creative representation $C_N$ of an unrelated video. The creative swap contrastive loss is formulated as:

\begin{equation}
\begin{aligned}
\mathcal{L}_{\mathrm{CSCL}}
&= -
\sum_{\widetilde{C}\in\mathcal{S}}
\sum_{p\in\mathcal{P}}
\log\frac{\exp(s(\widetilde{C},p)/\tau)}
{D(\widetilde{C})},\\
D(\widetilde{C})
&= \sum_{q\in\mathcal{P}}
\exp(s(\widetilde{C},q)/\tau)+ \exp(s(\widetilde{C},C_N)/\tau),
\end{aligned}
\end{equation}

\noindent where $\mathcal{S}\in \{\widetilde{C}_{A\rightarrow B},
\widetilde{C}_{B\rightarrow A}\}$ and $\mathcal{P}\in \{C_A,C_B\}$. $s(\cdot,\cdot)$ denotes
cosine similarity and $\tau$ is a temperature parameter.

\begin{figure}
  \centering
   \includegraphics[width=1.0\linewidth]{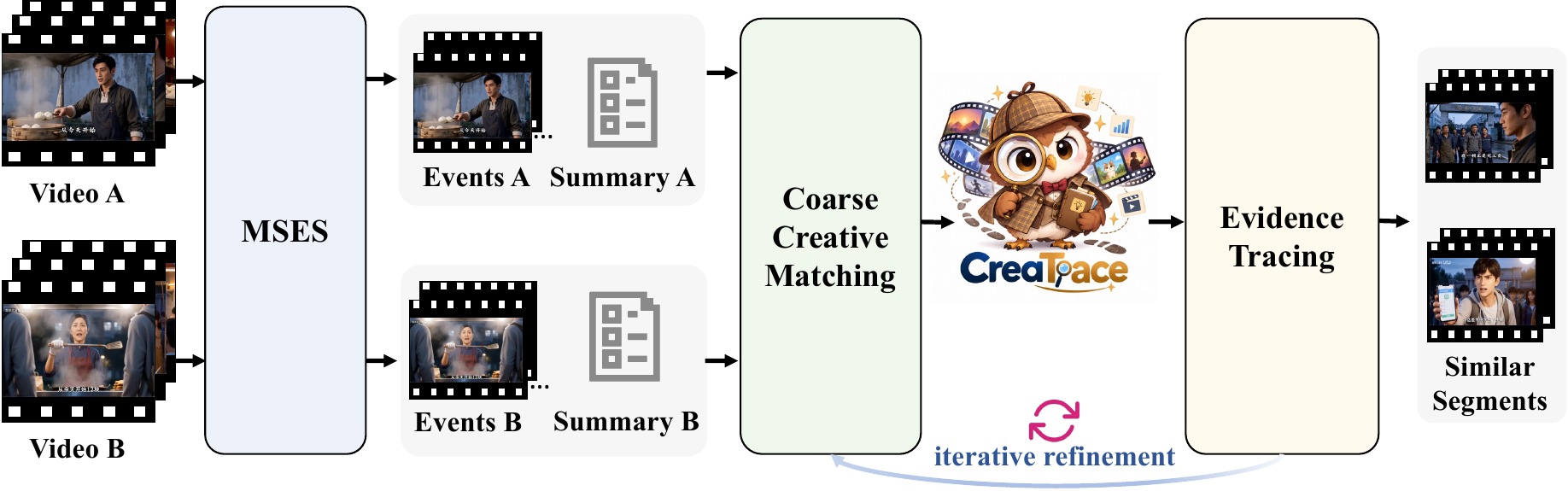}
\caption{Overview of our CreaTrace inference pipeline.}
   \label{overview}
\end{figure}
\begin{figure*}[t]
  \centering
   \includegraphics[width=1.0\linewidth]{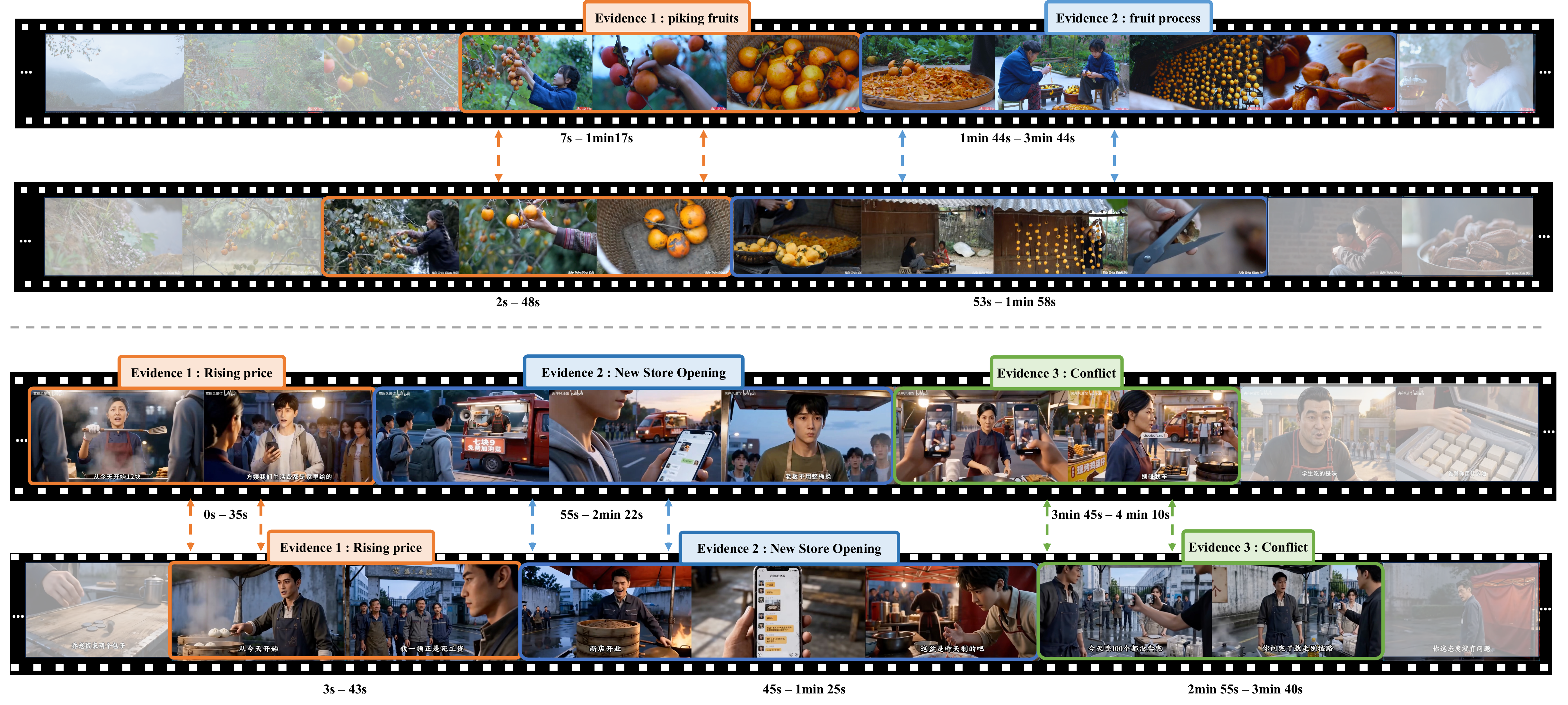}
   \vspace{-0.5cm}
   \caption{\textbf{Qualitative visualization of creative evidence localization by CreaTrace. }The two examples illustrate corresponding activity sequences and narrative structures across videos with different visual appearances. Colored boxes indicate localized evidence segments, while dashed arrows connect corresponding creative events across videos.}
   \label{result}
\end{figure*}
\noindent\textbf{Iterative Creative Evidence Tracing. }
Given a suspicious event pair, we divide each event into timestamped subclips and extract their local video features. At each iteration, creative queries interact with these video features to obtain creative representations for computing cross-video similarity, further to identify potential evidence pairs. Subsequently, the localized evidence is then fed back to update the creative queries, which are used to again extract creative representations and refine cross-video matching in the next iteration. This iterative process progressively narrows to highly similar subclip pairs, which are merged into temporal evidence intervals and used to generate interpretable reports for creative similarity verification.

\subsection{Training and Inference}
\noindent\textbf{Training.} We first apply MSES to divide videos into semantic events and perform Coarse Creative Matching to retrieve candidate event pairs. Based on these candidate event pairs, positive and negative event pairs are constructed to train Creative Decoupling and Creative Swap Contrastive Learning. During training, the pretrained video backbone remains frozen, while the decoupling modules and learnable queries are jointly optimized to disentangle underlying creative patterns from visual realizations, while enhancing creative representation consistency across videos.\par

\noindent\textbf{Inference.} As illustrated in Figure~\ref{overview}, MSES first organizes the input videos into semantic events, and coarse creative matching identifies potentially suspicious event pairs. Guided by the learned creative representations, CreaTrace iteratively traces and refines the corresponding evidence segments across both videos. The localized evidence is then fed back into the coarse matching stage to update creative similarity scores for detection. This feedback mechanism enables progressive evidence localization and creative similarity verification within a unified framework.\par

\newcommand{\gain}[1]{{\scriptsize\, (+#1\%)}}

\section{Experiments}
\subsection{Experiment Setups}
\textbf{Implementation details. }In Multimodal Semantic Event Segmentation, we combine WhisperX \cite{bain2023whisperx} speech transcription, PySceneDetect shot detection and Qwen3-VL-8B-Instruct \cite{bai2025qwen3} descriptions to obtain multimodal event representations. Event summaries and action sequences are encoded using bge-m3 \cite{chen2024m3} for coarse candidate matching. For CreaTrace, we extract features using a frozen V-JEPA 2.1 ViT-G encoder \cite{jepa}. Each temporal window contains 16 frames sampled at 4 FPS. CreaTrace uses eight learnable creative queries, a four-layer query encoder, a two-layer decoupling encoder and a two-layer decoupling decoder. The head of attention modules is set to 8.\par
\noindent\textbf{Baselines. }For detection, we compare CreaTrace with four representative open-source large multimodal models capable of video understanding: Qwen3-VL-8B-Instruct \cite{bai2025qwen3}, Qwen2.5-VL-7B-Instruct \cite{bai2025qwen25vl}, InternVL3-8B \cite{zhu2025internvl3} and VideoLLaMA3-7B \cite{zhang2025videollama}. All models are evaluated in a zero-shot setting without specific task fine-tuning. For evaluation, we adopt a standardized five-point scale to quantify creative similarity, with scores ranging from 1 (distinct creative mechanisms) to 5 (highly similar underlying creative designs). To account for the variations in score distributions of specific model, we determine the optimal decision threshold for each baseline by maximizing the F1-score on the validation set and keep it fixed during test. For CreaTrace, we average the top-3 highest event pair similarities to obtain the similarity score and use $\tau=0.70$ as the default decision threshold. For localization, we compare CreaTrace with three representative open-source video copy localization methods: TransVCL \cite{he2023transvcl}, RTR \cite{lu2024self} and VCSL-SPD \cite{vcsl}, which primarily target visually duplicated or transformed video content.

\definecolor{bestred}{RGB}{255,204,204}
\definecolor{secondyellow}{RGB}{255,247,204}

\begin{table*}[t]
\centering
\caption{
Quantitative comparison with zero-shot large multimodal models in video creative plagiarism detection. The best and second-best results are
highlighted in \colorbox{bestred}{red} and
\colorbox{secondyellow}{yellow}, respectively.
$\uparrow$ indicates that higher values are better.
}
\label{lmm_comparison}

\small
\renewcommand{\arraystretch}{1.12}

\begin{tabular}{l|ccccc}
\toprule[1.4pt]

Model
& ACC $\uparrow$
& AP $\uparrow$
& ROC-AUC $\uparrow$
& Recall@0.25 $\uparrow$
& Recall@0.5 $\uparrow$ \\

\midrule[0.6pt]

Qwen3-VL-8B-Instruct \cite{bai2025qwen3}
& \cellcolor{secondyellow}0.5250
& \cellcolor{secondyellow}0.6295
& \cellcolor{secondyellow}0.6925
& \cellcolor{secondyellow}0.3250
& \cellcolor{secondyellow}0.6250 \\

Qwen2.5-VL-7B-Instruct \cite{bai2025qwen25vl}
& 0.5000
& 0.5000
& 0.5000
& 0.2500
& 0.5000 \\

InternVL3-8B \cite{zhu2025internvl3}
& 0.5000
& 0.6250
& 0.6703
& \cellcolor{secondyellow}0.3250
& 0.6000 \\

VideoLLaMA3-7B \cite{zhang2025videollama}
& 0.5000
& 0.5479
& 0.5875
& 0.2750
& 0.5500 \\

\midrule[1.1pt]

\textbf{CreaTrace (Ours)}
& \cellcolor{bestred}0.6875
& \cellcolor{bestred}0.9048
& \cellcolor{bestred}0.8888
& \cellcolor{bestred}0.4750
& \cellcolor{bestred}0.8250 \\

\bottomrule[1.4pt]
\end{tabular}
\end{table*}

\begin{table*}[t]
\centering
\caption{
Quantitative comparison with existing video copy localization methods in creative evidence localization. P, R, and F1 denote Precision, Recall, and F1-score, respectively, measured at tIoU thresholds of 0.1, 0.3, and 0.5.
The best and second-best results are highlighted in
\colorbox{bestred}{red} and
\colorbox{secondyellow}{yellow}, respectively.
$\uparrow$ indicates that higher values are better.
}
\label{localization_comparison}

\footnotesize
\setlength{\tabcolsep}{5pt}
\renewcommand{\arraystretch}{1.12}

\begin{tabular}{l|ccccccccc}
\toprule[1.4pt]

Method
& P@0.1 $\uparrow$
& R@0.1 $\uparrow$
& F1@0.1 $\uparrow$
& P@0.3 $\uparrow$
& R@0.3 $\uparrow$
& F1@0.3 $\uparrow$
& P@0.5 $\uparrow$
& R@0.5 $\uparrow$
& F1@0.5 $\uparrow$ \\

\midrule[0.6pt]

TransVCL \cite{he2023transvcl}
& 13.64
& \cellcolor{secondyellow}12.50
& 13.04
& 4.55
& 4.17
& 4.35
& 0.00
& 0.00
& 0.00 \\

RTR \cite{lu2024self}
& \cellcolor{secondyellow}15.00
& \cellcolor{secondyellow}12.50
& \cellcolor{secondyellow}13.64
& \cellcolor{secondyellow}10.00
& \cellcolor{secondyellow}8.33
& \cellcolor{secondyellow}9.09
& \cellcolor{secondyellow}5.00
& \cellcolor{secondyellow}4.17
& \cellcolor{secondyellow}4.55 \\

VCSL-SPD \cite{vcsl}
& 11.11
& 8.33
& 9.52
& 5.56
& 4.17
& 4.76
& 0.00
& 0.00
& 0.00 \\

\midrule[1.1pt]

\textbf{CreaTrace (Ours)}
& \cellcolor{bestred}62.07
& \cellcolor{bestred}75.00
& \cellcolor{bestred}67.92
& \cellcolor{bestred}48.28
& \cellcolor{bestred}58.33
& \cellcolor{bestred}52.83
& \cellcolor{bestred}27.59
& \cellcolor{bestred}33.33
& \cellcolor{bestred}30.19 \\

\bottomrule[1.4pt]
\end{tabular}
\end{table*}

















\noindent\textbf{Evaluation Metric.} We evaluate all methods from both classification and localization perspectives. For classification, we report Accuracy (ACC), Average Precision (AP) and Area Under the Receiver Operating Characteristic Curve (ROC-AUC) to assess overall prediction accuracy and discriminative performance. Additionally, we further report recall using Recall@0.25 and Recall@0.5, measuring the proportion of ground-truth positive video pairs retrieved within the top 25\% and 50\% of all candidate pairs ranked by similarity scores, respectively. For localization, we measure temporal Intersection over Union (tIoU) between predicted and ground-truth intervals. Since each instance involves two corresponding video segments, we define pairwise tIoU as the minimum tIoU across both videos. Meanwhile, we report segment-level Precision, Recall, and F1-score at tIoU thresholds of 0.1, 0.3 and 0.5 to assess evidence localization performance under varying overlaps.\par

\noindent\textbf{Datasets. }We collect 40 publicly available videos from 10 creative groups, totaling 17.82 hours, where videos within each group share common creative patterns. Through MSES, we obtain 515 semantic events with an average duration of 124.57 seconds. Based on group membership, we construct 22,137 candidate positive event pairs from videos within the same group and 103,299 negative event pairs from videos belonging to different groups. To prevent data leakage, we split the data at the creative-group level, ensuring that no creative group appears in both the training and test sets. We select 20,000 informative event pairs from the training groups using coarse creative matching, and construct a balanced test set of 80 video pairs from the test groups, including 40 positive and 40 negative pairs.

\subsection{Quantitative and Qualitative Results}
\noindent\textbf{Quantitative Results. }For detection, Table~\ref{lmm_comparison} compares our CreaTrace with representative zero-shot MLLMs on creative plagiarism detection, which shows our CreaTrace consistently outperforms all baselines across all evaluation metrics, achieving substantial improvements in ACC, AP and ROC-AUC. Although existing MLLMs exhibit strong video understanding capabilities, they struggle to extract underlying creative patterns from specific visual content, making it difficult to identify creative similarities. In contrast, our CreaTrace explicitly disentangles abstract creative patterns from visual content, enabling more accurate and reliable creative plagiarism detection.\par
\renewcommand{\arraystretch}{1.5}
\begin{table*}[t]
\centering
\caption{
\textbf{Quantitative comparison of creative plagiarism detection
performance across different MLLMs using CreaTrace's
iteratively localized evidence.}
Percentages indicate relative improvements over the original
video inputs. The best and second-best results are highlighted in
\colorbox{bestred}{red} and
\colorbox{secondyellow}{yellow}, respectively.
}
\label{iteration_comparison}

\small
\renewcommand{\arraystretch}{1.15}
\setlength{\tabcolsep}{4pt}

\resizebox{\textwidth}{!}{
\begin{tabular}{l|ccccc|ccccc}
\toprule[1.5pt]

\multirow{2}{*}{Model}
& \multicolumn{5}{c|}{Iteration 1}
& \multicolumn{5}{c}{Iteration 2} \\

\cmidrule(lr){2-6}
\cmidrule(lr){7-11}

& ACC $\uparrow$
& AP $\uparrow$
& ROC-AUC $\uparrow$
& Recall@0.25 $\uparrow$
& Recall@0.5 $\uparrow$
& ACC $\uparrow$
& AP $\uparrow$
& ROC-AUC $\uparrow$
& Recall@0.25 $\uparrow$
& Recall@0.5 $\uparrow$ \\

\midrule[0.8pt]

Qwen3-VL-8B-Instruct
& \cellcolor{bestred}0.6500\gain{23.81}
& \cellcolor{bestred}0.8107\gain{28.78}
& \cellcolor{bestred}0.8019\gain{15.80}
& \cellcolor{bestred}0.4250\gain{30.77}
& \cellcolor{bestred}0.7750\gain{24.00}
& \cellcolor{bestred}0.6625\gain{26.19}
& \cellcolor{bestred}0.8354\gain{32.71}
& \cellcolor{bestred}0.8263\gain{19.32}
& \cellcolor{bestred}0.4500\gain{38.46}
& \cellcolor{bestred}0.8000\gain{28.00} \\

Qwen2.5-VL-7B-Instruct
& 0.5750\gain{15.00}
& 0.6704\gain{34.08}
& 0.6694\gain{33.88}
& 0.3500\gain{40.00}
& 0.6500\gain{30.00}
& 0.6000\gain{20.00}
& 0.7103\gain{42.06}
& 0.7050\gain{41.00}
& 0.3750\gain{50.00}
& 0.6750\gain{35.00} \\

InternVL3-8B
& \cellcolor{secondyellow}0.6250\gain{25.00}
& \cellcolor{secondyellow}0.7706\gain{23.30}
& \cellcolor{secondyellow}0.7594\gain{13.29}
& \cellcolor{secondyellow}0.4000\gain{23.08}
& \cellcolor{secondyellow}0.7250\gain{20.83}
& \cellcolor{secondyellow}0.6500\gain{30.00}
& \cellcolor{secondyellow}0.8018\gain{28.29}
& \cellcolor{secondyellow}0.7881\gain{17.57}
& \cellcolor{secondyellow}0.4250\gain{30.77}
& \cellcolor{secondyellow}0.7500\gain{25.00} \\

VideoLLaMA3-7B
& 0.6000\gain{20.00}
& 0.7197\gain{31.36}
& 0.7075\gain{20.43}
& 0.3750\gain{36.36}
& 0.6750\gain{22.73}
& 0.6250\gain{25.00}
& 0.7543\gain{37.67}
& 0.7425\gain{26.38}
& 0.4000\gain{45.45}
& 0.7000\gain{27.27} \\

\bottomrule[1.5pt]
\end{tabular}
}
\end{table*}
For evidence localization, as shown in Table \ref{localization_comparison}, CreaTrace achieves superior performance over existing video copy localization methods across different tIoU thresholds, demonstrating its effectiveness in accurately identifying corresponding evidence segments. Traditional methods primarily rely on visual content matching, limiting their ability to localize creative evidence across visually distinct videos. In contrast, CreaTrace effectively captures underlying creative patterns and traces corresponding temporal evidence.\par

\noindent\textbf{Qualitative Results. }Figure~\ref{result} presents the creative evidence localization examples by our CreaTrace, demonstrating its ability to identify corresponding temporal segments associated with similar creative events across different videos. In the first example, CreaTrace identifies a shared creative progression from fruit picking to subsequent processing, revealing consistent event organization across different visual realizations. In the second example, both videos follow a similar narrative structure involving price increases, new store openings and subsequent conflicts. Above results illustrate our CreaTrace can trace and localize supporting evidence for creative similarity assessment. More qualitative creative evidence localization results are presented in the \textbf{Supplementary Material}.\par

\begin{table}[t]
\centering
\caption{
\textbf{Ablation study of key designs in CreaTrace.} CSCL denotes Creative Swap Contrastive Learning.
}
\label{ablation}

\footnotesize
\setlength{\tabcolsep}{3pt}
\renewcommand{\arraystretch}{1.12}

\resizebox{\columnwidth}{!}{
\begin{tabular}{lcccc}
\toprule[1.4pt]

Variant
& ACC $\uparrow$
& AP $\uparrow$
& P@0.3 $\uparrow$
& F1@0.3 $\uparrow$ \\

\midrule[0.6pt]

Baseline
& 0.5625
& 0.6824
& 10.34
& 11.32 \\

+ Creative Decoupling
& \cellcolor{secondyellow}0.6125
& \cellcolor{secondyellow}0.7586
& \cellcolor{secondyellow}27.59
& \cellcolor{secondyellow}30.19 \\

+ CSCL (Full)
& \cellcolor{bestred}0.6875
& \cellcolor{bestred}0.9048
& \cellcolor{bestred}48.28
& \cellcolor{bestred}52.83 \\

\bottomrule[1.4pt]
\end{tabular}
}
\end{table}

\subsection{Ablation Studies}

\begin{figure}
  \centering
   \includegraphics[width=1.0\linewidth]{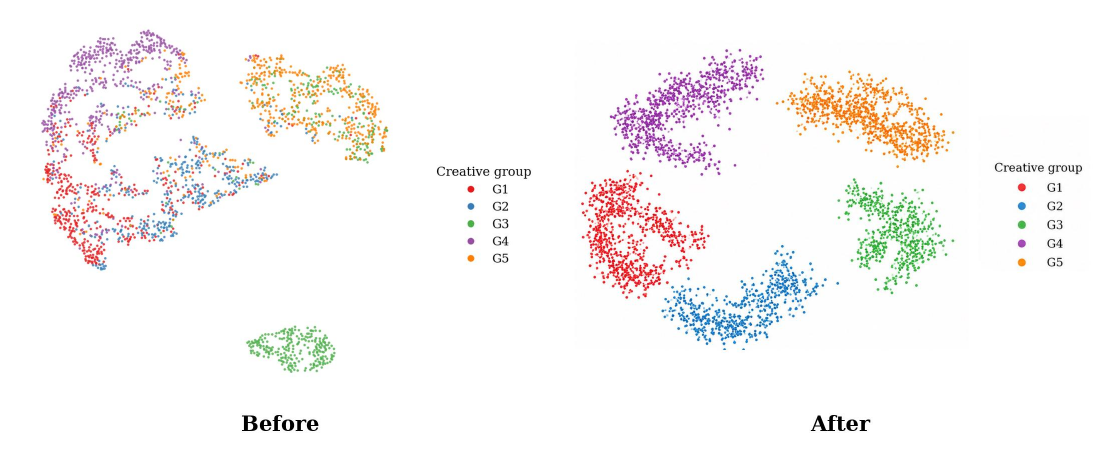}
\caption{t-SNE visualization of creative representations before and after creative disentanglement.}
   \label{tsne}
\end{figure}

\textbf{Effectiveness of Creative Decoupling. }As shown in Table~\ref{ablation}, incorporating Creative Decoupling improves creative similarity detection, demonstrating its effectiveness in capturing shared creative semantics across different visual realizations. Furthermore, the t-SNE visualization in Figure~\ref{tsne} shows that similar creative semantics tend to form more compact clusters, further validating the effectiveness of Creative Decoupling in learning creative representations.

\noindent\textbf{Effectiveness of Creative Swap Contrastive Learning. }As shown in Table~\ref{ablation}, introducing Creative Swap Contrastive Learning improves both detection and localization performance, demonstrating its effectiveness in learning discriminative creative representations. Furthermore, the loss ablation results in Table~\ref{loss} confirm the importance of swap contrastive supervision in preserving creative consistency.

\noindent\textbf{Sensitivity of the Coarse Creative Matching Threshold $\tau$.}
As shown in Table~\ref{threshold_sensitivity}, increasing $\tau$ generally improves precision while reducing recall, reflecting a trade-off between false matches and missed candidates. Notably, $\tau=0.70$ achieves the highest ACC and F1 while maintaining the second-highest recall, effectively balancing candidate selection accuracy and retrieval coverage. We therefore adopt $\tau=0.70$ as the default threshold.

\begin{table}[htbp]
\centering
\caption{
Sensitivity analysis of the threshold $\tau$.
}
\label{threshold_sensitivity}

\footnotesize
\renewcommand{\arraystretch}{1.18}

\begin{tabular*}{0.82\columnwidth}{
@{\extracolsep{\fill}}c|cccc
}
\toprule[1.2pt]

$\tau$
& ACC $\uparrow$
& Precision $\uparrow$
& Recall $\uparrow$
& F1 $\uparrow$ \\

\midrule[0.6pt]

0.60
& 0.5875
& 0.5479
& \cellcolor{bestred}1.0000
& 0.7080 \\

0.65
& 0.6000
& 0.5556
& \cellcolor{bestred}1.0000
& 0.7143 \\

0.70
& \cellcolor{bestred}0.6875
& 0.6230
& \cellcolor{secondyellow}0.9500
& \cellcolor{bestred}0.7525 \\

0.75
& \cellcolor{secondyellow}0.6750
& \cellcolor{secondyellow}0.6667
& 0.7000
& 0.6829 \\

0.80
& 0.6125
& \cellcolor{bestred}1.0000
& 0.2250
& 0.3673 \\

\bottomrule[1.2pt]
\end{tabular*}
\end{table}












\begin{table}[htbp]
\centering
\caption{
Ablation study of the loss functions in CreaTrace.
}
\label{loss}

\footnotesize
\renewcommand{\arraystretch}{1.15}

\begin{tabular*}{0.9\columnwidth}{
@{\extracolsep{\fill}}cc|cccc}
\toprule[1.2pt]

$\mathcal{L}_{crea}$
& $\mathcal{L}_{CSCL}$
& ACC $\uparrow$
& AP $\uparrow$
& P@0.3 $\uparrow$
& F1@0.3 $\uparrow$ \\

\midrule[0.6pt]

$\checkmark$
& --
& 0.6125
& 0.7586
& 27.59
& 30.19 \\

--
& $\checkmark$
& \cellcolor{secondyellow}0.6500
& \cellcolor{secondyellow}0.8260
& \cellcolor{secondyellow}37.93
& \cellcolor{secondyellow}41.51 \\

$\checkmark$
& $\checkmark$
& \cellcolor{bestred}0.6875
& \cellcolor{bestred}0.9048
& \cellcolor{bestred}48.28
& \cellcolor{bestred}52.83 \\

\bottomrule[1.2pt]
\end{tabular*}
\end{table}

\subsection{Generalizability}
To evaluate the effectiveness of creative evidence localized by CreaTrace, we investigate whether it can improve creative plagiarism detection across different MLLMs, including Qwen3-VL \cite{bai2025qwen3}, Qwen2.5-VL \cite{bai2025qwen25vl}, InternVL3 \cite{zhu2025internvl3} and VideoLLaMA3 \cite{zhang2025videollama}. As shown in Table~\ref{iteration_comparison}, replacing the original video inputs with localized evidence consistently improves detection performance across all four models, demonstrating the effectiveness of progressive evidence refinement. Above results highlight the generalizability of creative evidence localized by our CreaTrace across different MLLMs.\par
\section{Conclusion}
In this paper, we introduce \textbf{Video Creative Plagiarism Detection}, a novel task that aims to identify shared creative patterns across visually different videos and localize the corresponding evidence. To address this challenge, we first develop a Multimodal Semantic Event Segmentation strategy that organizes long videos into semantically coherent events, constructing an event-level dataset for fine-grained creative similarity analysis. Furthermore, we propose \textbf{CreaTrace}, an iterative creative evidence tracing framework that disentangles underlying creative patterns from diverse visual realizations through Creative Decoupling and Creative Swap Contrastive Learning. By iteratively refining creative similarity assessments and tracing supporting evidence, CreaTrace enables more precise video creative plagiarism detection and localization while supporting explainable evidence report. Extensive experiments demonstrate the effectiveness of CreaTrace across diverse creative plagiarism scenarios. We hope our work will provide new insights into video creative plagiarism detection and encourage further exploration of this emerging research direction.\par
{
    \small
    \bibliographystyle{ieeenat_fullname}
    \bibliography{main}
}


\end{document}